\documentclass[11pt]{article}
\usepackage{coling2020}
\usepackage{times}
\usepackage{url}
\usepackage{latexsym}

\usepackage{algorithm}
\usepackage{algpseudocode}
\usepackage[table]{xcolor}
\usepackage{enumitem}
\usepackage{subcaption}
\usepackage{multirow}
\usepackage{graphicx}
\usepackage{tikz}
\usepackage{collcell}
\usepackage{amssymb}
\usepackage{stackengine}
\usepackage{booktabs}
\usepackage{bm}
\usepackage{verbatim}
\usepackage[font=small,labelfont=bf]{caption}

\usepackage[hidelinks]{hyperref}
\usepackage{xcolor,soul,lipsum}
\newcommand{\myul}[2][black]{\setulcolor{#1}\ul{#2}\setulcolor{black}}

\newcommand*{\MinNumber}{0}
\newcommand*{\MaxNumber}{1}
\newcommand{\ApplyGradient}[1]{%
        \pgfmathsetmacro{\PercentColor}{70.0*(#1-\MinNumber)/(\MaxNumber-\MinNumber)}
        \hspace{-0.33em}\colorbox{green!\PercentColor!white}{}
}
\newcolumntype{R}{>{\collectcell\ApplyGradient}c<{\endcollectcell}}

\colingfinalcopy 

\title{Incorporating Temporal Information in Entailment Graph Mining}

\author{Liane Guillou$^{\dagger*}$, Sander Bijl De Vroe$^{\dagger*}$, Mohammad Javad Hosseini$^{\dagger}$$^{\ddagger}$,\\ \textbf{Mark Johnson$^{\mathsection}$, and Mark Steedman$^{\dagger}$} \\
  $^{\dagger}$University of Edinburgh, $^{\ddagger}$ The Alan Turing Institute, UK, $^{\mathsection}$Macquarie University \\
  {\tt liane.guillou@ed.ac.uk, sbdv@ed.ac.uk, javad.hosseini@ed.ac.uk}\\
 {\tt mark.johnson@mq.edu.au, steedman@inf.ed.ac.uk}}
\date{}

\begin{document}
\maketitle

%
\blfootnote{
   \hspace{-0.65cm}  
    This work is licensed under a Creative Commons Attribution 4.0 International Licence. Licence details: \url{http://creativecommons.org/licenses/by/4.0/}.
}
\begingroup\renewcommand\thefootnote{*}
\footnotetext{The first two authors contributed equally to this work}
\endgroup

\begin{abstract}
We present a novel method for injecting temporality into entailment graphs to address the problem of spurious entailments, which may arise from similar but temporally distinct events involving the same pair of entities. We focus on the sports domain in which the same pairs of teams play on different occasions, with different outcomes. We present an unsupervised model that aims to learn entailments such as win/lose $\rightarrow$ play, while avoiding the pitfall of learning non-entailments such as win $\not\rightarrow$ lose. We evaluate our model on a manually constructed dataset, showing that incorporating time intervals and applying a temporal window around them, are effective strategies.
\end{abstract}

\section{Introduction}
\label{sec:introduction}

Recognising textual entailment and paraphrases is core to many downstream NLP applications such as question answering and semantic parsing. In the case of open-domain question answering over unstructured data, the answer to a question may not be explicitly stated in the text, but may be recovered via paraphrases and/or entailment rules.
 
Entailment graphs \cite{berant2011,berant2015,hosseini2018}, in which nodes represent predicates and edges are entailment relations, have been proposed as a means to answer such questions. They can be mined using unsupervised methods applied over large collections of text, by keeping track of which entity pairs occur with which predicates. One common error made by these graphs, however, is that they assert spurious associations between similar but temporally distinct events that occur with the same entity pairs. For example, both the predicates \textit{beat} and \textit{lost against} will apply to sports team entity pairs such as (\textbf{Arsenal}, \textbf{Man United}). This is likely to mislead the current methods into incorrectly assigning an entailment relation between these two predicates.

\begin{figure}[h!]
\begin{subfigure}[]{.7\linewidth}
\begin{tabular}{l}
\textbf{Arsenal}-\textit{played} and \textit{lost against}-\textbf{Man United} 1-3 (25/01/2019)\\
\textbf{Arsenal}-\textit{played} and \textit{beat}-\textbf{Man United} 2-0 (10/03/2018)\\
\textbf{Arsenal}-\textit{played} and \textit{tied with}-\textbf{Man United} 1-1 (30/09/2019) \\
\end{tabular}
\end{subfigure}%
\begin{subfigure}[]{.3\linewidth}
\centering
\includegraphics[scale=0.6]{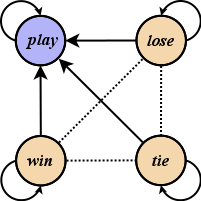}
\label{fig:example-entailments}
\end{subfigure}
\caption{Example sentences (left) and their resulting (collapsed) entailment/non-entailment graph (right)}
\label{fig:example}
\end{figure}

In this paper we extend the framework of \newcite{hosseini2018} to incorporate the temporal location of events, with the aim of mitigating these spurious entailments. Temporal information can be used to disentangle these groups of highly correlated predicates, because although they will share entity pairs, they will never occur at the same time. For example, in Figure~\ref{fig:example} Arsenal and Man United played each other three times in 2019, with three different outcomes: \textbf{win} (\textit{beat}), \textbf{lose} (\textit{lost against}), \textbf{tie} (\textit{tied with}).

In previous methods, the context in which the predicates occur appears to be the same, because they only consider entity pairs as context. Therefore they mistakenly take the examples in Figure~\ref{fig:example} as evidence of entailments or paraphrases between the three outcome predicates (win, lose, and tie), depending on the distributions found in the data. Our method enriches this context to include time interval information, thereby filtering out combinations that are not temporally near each other. Thus we hope to avoid learning that \textit{beat} $\rightarrow$ \textit{lost against}, while still learning that \textit{beat} $\rightarrow$ \textit{play}.
 
As an initial test domain, we focus on the sports news genre, using extracted relations that involve two sports teams. We evaluate on a manually constructed dataset of 1,312 entailment pairs based on paraphrases of the predicates in the graph on the right hand side of Figure~\ref{fig:example}. Our goal is to recover the structure of this graph in an unsupervised way, separating each of the highly correlated outcome predicates while predicting that they all entail \textit{play}.
 
The contributions of this work are: 1) a model for incorporating relation-level time intervals into an entailment graph mining procedure, outperforming non-temporal models, and 2) a manually constructed evaluation dataset of sports domain predicates. To our knowledge this is the first attempt to incorporate temporal information for learning entailment graphs.

\section{Related Work}
\label{sec:related_work}

\subsection{Entailment Graphs}
\label{subsec:entailment_graphs}

Entailment graphs have been constructed for a range of domains, including newswire \cite{hosseini2018}, health \cite{levy2014}, and commonsense \cite{yu2020}. In order to leverage temporal information, our work focuses on the news domain, in which each article has a known publication date and temporal expressions are commonly used.

A range of node representations have been explored, including Open-IE propositions \cite{levy2014}, typed predicates \cite{berant2011,hosseini2018}, and eventualities \cite{yu2020}. In this work we use typed predicates, leveraging the second level in the FIGER hierarchy \cite{ling2012}, to enable a close examination of events that take place between two sports teams.

Whether predicates in the graph entail each other may be determined using a variety of similarity measures. These are inspired by the Distributional Inclusion Hypothesis, which states that a predicate \textit{p} entails another predicate \textit{q} if for any context in which \textit{p} can be used, \textit{q} may be used in its place \cite{dagan1999,geffet2005}. They include the symmetric Lin's similarity measure \cite{lin1998}, the directional Weeds' precision and recall measures \cite{weeds2003}, and the Balanced Inclusion score (BInc) \cite{szpektor2008}. BInc, the geometric mean of Lin’s similarity and Weed’s precision, combines the desirable behaviors of symmetric and directional measures. We adapt and examine each of these similarity measures using our evaluation dataset.

Alternatively, \newcite{hosseini2019} performed link prediction on the set of relation triples extracted from the text, and showed improvements over BInc by augmenting the data with additional predicted triples. We consider this link prediction model to be beyond the scope of this work.

\subsection{Evaluating Entailment Graphs}
\label{subsec:evaluating_entailment_graphs}

The construction of entailment datasets has been framed as a number of manual annotation tasks including image captioning \cite{bowman2015} and question answering \cite{levy2016}.

The dataset creation method used by \newcite{levy2016} aims to address the bias towards real world knowledge. They ask human annotators to mark possible answers to questions as True/False (entailment/non-entailment), with entities in the answer replaced using tokens representing their type (e.g. \textit{London} becomes \textit{city}). The method also aims to address the bias of other datasets such as Zeichner's dataset \cite{zeichner2012} and the SherLIic dataset \cite{schmitt2019}, in which candidate entailments were automatically pre-selected for manual annotation according to a similarity measure. Entailments that exist, but are not captured by these similarity measures will therefore be excluded.

There has been very little work on the specific problem of evaluating entailment of a temporal nature. The FraCas test suite \cite{cooper1996} contains a small section of which only a few examples are entailments between predicates. The TEA dataset \cite{kober2019} consists of pairs of sentences in which temporally ordered predicates have varying tense and aspect, such as \textit{is visiting} $\rightarrow$ \textit{has arrived}, but does not include non-entailments that can be learned through the temporal separation of events (such as the outcome predicates \textit{win} and \textit{lose} that we are interested in). Since there is no dataset to evaluate this phenomenon, we construct our own (Section~\ref{subsec:dataset}). Our method for dataset construction is similar to that of \newcite{berant2011}. They manually annotated all edges in 10 typed entailment graphs, resulting in 3,427 edges (entailments) and 35,585 non-edges (non-entailments).

\section{Method}
\label{sec:method}

\subsection{Relation Extraction}
\label{subsec:relation_extraction}
We use a pipeline based on a Combinatory Categorial Grammar (CCG, \cite{Steedman2000}) parser to extract binary relations with time intervals. These relations are used to construct typed entailment graphs using the unsupervised method of \newcite{hosseini2018}, adapted to compare only pairs of relations that are temporally near each other. We extract binary relations of the form \textbf{arg1}-\textit{predicate}-\textbf{arg2} (e.g. \textbf{Arsenal}-\textit{beat}-\textbf{Man United}), following the example of \newcite{lewis2013} and \newcite{berant2015}. We use a pipeline approach similar to that described by \newcite{hosseini2018}, which allows us to extract open-domain relations. Relations are extracted from the NewsSpike corpus \cite{zhang2013} of news articles collected from multiple sources over a period of approximately six weeks.

We traverse dependency graphs generated over the output of the Rotating CCG parser \cite{stanojevic2019}, starting from verb and preposition nodes, until we reach an argument leaf node. The traversed nodes are used to form (lemmatised) predicate strings, and arguments are classified as either a Named Entity (extracted by the CoreNLP Named Entity recogniser), or a general entity (all other nouns and noun phrases). Predicate strings may include (non-auxiliary) verbs, verb particles, adjectives, and prepositions. Negation nodes are detected via string match (``not", ``n't", and ``never"), and are included in the predicate if there is a path between the negation node and a node in the predicate. We map passive predicates to active ones. Modifiers such as ``managed to" as in the example ``Arsenal managed to beat Man United" are also extracted and included in the predicate. As the modifiers may be rather sparse, we extract the relation both with and without the modifier.

We extract and resolve time expressions in the document text, using SUTime \cite{chang2012}, available via CoreNLP. If there is a path in the CCG dependency graph between the time expression and a node in the predicate, the relation is assigned a time interval. Entities are mapped to types by linking to their Freebase \cite{bollacker2008} IDs using AIDA-Light \cite{nguyen2014}, and subsequently mapping these IDs to their fine-grained FIGER types \cite{ling2012}.

To restrict the data to the sports domain we filter the set of output relations, accepting only those involving two entities of the fine-grained FIGER type \textit{organization/sports\_team}. This results in a set of 78,439 binary relations extracted from 24,147 articles, of which 14,664 (approximately 19\%) have time intervals derived from SUTime. The sports domain has the advantage that events are similar and should be easily separable in time, and it provides the straightforward \textit{win}-\textit{lose}-\textit{tie} outcome set. Sports data is common in NewsSpike, and sports teams have reliable Named Entity linking, making it suitable for an initial investigation. In the future we will apply this method to other entity type pairs.

\subsection{Graph Construction}
\label{subsec:graph_construction}

The input to the graph construction step is the set of typed binary relations paired with their time intervals. As we focus on events that involve two sports teams, the output is a single \textbf{organization}-\textbf{organization} graph, rather than the typical set of graphs (one for every pair of types). Note that these graphs contain only locally learned entailments, and that global inference across graphs is not performed. This is sufficient to demonstrate the benefit of incorporating time intervals.

In the original method for computing local entailment scores, \newcite{hosseini2018} extract a feature vector for each typed predicate (e.g. \textit{play} with type pair organization-organization). The entity pairs from the binary relations (e.g. \textbf{Arsenal}, \textbf{Man United}) are used as the feature types, and the pointwise mutual information (PMI) between the predicate and the entity pair is the value. These feature vectors are then used to compute local similarity scores. 

We extend this method to consider the time intervals for each of the binary relations, with the goal of comparing only those events that are temporally near each other. To achieve this, we filter the counts of predicate \textit{q} according to whether each event's time interval overlaps with any of \textit{p}'s. In other words, an event in \textit{q} is retained if it is close enough to any event in \textit{p}. We consider new local similarity scores based on both the filtered counts, and scaled PMI scores. 

Algorithm~\ref{temporal_filtering_2} describes the process of filtering counts using time intervals. The process uses a set of edges $\mathcal{E}$ between predicate nodes to store filtered count information. We loop through each entity pair \textit{ep} and get the list of predicates that occur with that entity pair (line 4). Then, for each pair of predicates, we instantiate \textit{edgeObjects} (line 7) between predicates \textit{p} and \textit{q} (in both directions), to store the filtered count information. We also retrieve \textit{p} and \textit{q}'s \textit{timeObjects}, containing a list of the time intervals at which the predicate and entity pair co-occurred (lines 8--9). For each pair of time intervals we compute whether there is an overlap (lines 12--19). The filtered count is the total number of events in predicate \textit{p} that temporally overlap with any event in predicate \textit{q}. 

The count is stored in the \textit{edgeObject} $edge_{p,q}$. Once all counts have been collected, they are used to compute the similarity measures. The computation of temporal measures is identical to that of the non-temporal counterparts, but they use the filtered counts as input instead of the regular counts. Each \textit{edgeObject} populates a cell in $\mathbf{W}$, the sparse matrix of all similarity scores between predicates \textit{p} and \textit{q}, as presented by \newcite{hosseini2018}. The filtered counts are also used to scale the PMI scores (see section \ref{subsec:similarity_measures}).

Consider the following minimal worked example: Two matches between Arsenal and Man United, one where Arsenal wins (on 10/03/2018), and one where they lose (on 25/01/2019). This initially results in the following extracted predicates and counts: \textit{play} (2), \textit{win} (1), \textit{lose} (1). After filtering, we add a count of 1 to both the \textit{win} $\rightarrow$ \textit{play} and \textit{lose} $\rightarrow$ \textit{play} edges, and a count of 0 to the \textit{win} $\rightarrow$ \textit{lose} edge (and its reverse).

We consider three possible sources of time intervals: 1) the resolved time expressions extracted from raw text using SUTime, 2) the document creation date (provided as metadata in the NewsSpike corpus), and 3) a combination of the two -- using resolved time expressions where these are available, and backing off to the document creation date where they are not. The intuition behind using time expressions extracted from the article text is that these ought to more accurately pinpoint the time interval of the events. However, as such expressions may be sparse, we also investigate the use of the document creation date, under the assumption that sports news is likely to be reported very close the day of the event.

We also consider a temporal window to extend the time intervals by \textit{N} days either side. This would mitigate the problem of sports events being reported several days after the event, especially when we fall back to the document creation date. For sports matches we would expect to see a benefit in using a small window of a few days, and a detrimental effect as that window grows increasingly larger. Specifically, we expect that larger windows would render temporal information useless, preventing our model from being able to distinguish between two different matches involving the same pair of teams. Time interval source and window size are event-specific parameters that we experiment with in Section~\ref{sec:experiments}.

\begin{algorithm}[t]
  \small
  \caption{Temporal filtering in local graph computation}\label{temporal_filtering_2}
  \begin{algorithmic}[1]
    \Procedure{TemporalFilter}{$entity\_pairs,predicates$}
      \State $\mathcal{E}\gets initialiseAllEdgeObjects(predicates)$ \Comment{Initialise set of edges}
      \For{ep \textbf{in} entity\_pairs}
        \State $predicates_{ep}\gets getPredicatesForEntityPair(predicates,ep)$
        \For{$p \gets 0$ to $length(predicates_{ep})$}
            \For{$q \gets p+1$ to $length(predicates_{ep})$}
                \State $edge_{p,q}, edge_{q,p}\gets getEdgeObjects(\mathcal{E},p,q)$
                \State $time\_objects_{ep,p}\gets getTimeObjects(ep,p)$
                \State $time\_objects_{ep,q}\gets getTimeObjects(ep,q)$
                \State $overlap_{p}\gets initialiseVectorOfZeros(length(time\_objects_{ep,p}))$
                \State $overlap_{q}\gets initialiseVectorOfZeros(length(time\_objects_{ep,q}))$
                \For{$i \gets 0$ to $length(time\_objects_{ep,p})$}
                    \For{$j \gets 0$ to $length(time\_objects_{ep,q})$}
                      \If{$compareIntervals(time\_objects_{ep,p}[i],time\_objects_{ep,q}[j]) = 1$} 
                            \State $overlap_{p}.set(i,1)$
							\State $overlap_{q}.set(j,1)$
                        \EndIf
                    \EndFor
                \EndFor
                \State $edge_{p,q}.addEdgeCounts(sum(overlap_{p}))$\Comment{Events in \textit{p} that temporally overlap with any \textit{q}}
                \State $edge_{q,p}.addEdgeCounts(sum(overlap_{q}))$
            \EndFor
            \State $\mathcal{E}.update(edge_{p,q}, edge_{q,p})$
        \EndFor
      \EndFor
      \State \textbf{return} $\mathcal{E}$
    \EndProcedure
  \end{algorithmic}
\end{algorithm}

\section{Evaluation}
\label{sec:evaluation}

\subsection{Dataset Construction}
\label{subsec:dataset}

We propose a semi-automatic method to construct a small evaluation dataset based on manually constructed paraphrase clusters. We start with a small set of predicates for which we know the entailment pattern, in our case $\{win, play, lose$ and $tie\}$. We restrict the dataset to include only those binary relations that involve two sports teams, by filtering on the fine-grained FIGER \cite{ling2012} type \textit{organization/sports\_team}. We then order the predicates by their frequency in the corpus, and manually select paraphrases of our small set with a count of at least 20 (the 235 most frequent predicates). This results in four clusters of paraphrases, with sizes of 26, 8, 3 and 5 respectively for \textit{win}, \textit{lose}, \textit{tie} and \textit{play}. We then automatically generate entailment pairs (1,312 in total), labelling them according to the pattern in Table~\ref{tab:evaldataset}, with premises in the rows and hypotheses in the columns.

We include the \textit{paraphrase} category for completeness, although we are more interested in the effect of separating temporally disjoint sports match outcomes. The paraphrase category contains predicates of varying gradation, such as \textit{crush} suggesting a strong victory or \textit{eliminate} indicating that a team is knocked out of a tournament. We wish to avoid specific predicates such as \textit{eliminate} entailing non-specific predicates like \textit{beat}. To avoid this issue we manually annotated the predicates for specificity, and for the \textit{paraphrase} entailments subset we only generate pairs for non-specific predicates. More generally, a set of paraphrase clusters with a total of $n$ predicates yields $n^2 - n$ pairs (not taking into account the paraphrase subsets reduction).\footnote{The subtracted term comes from duplicate pairs like \textit{defeat}-\textit{defeat}} The dataset of 1,312 entailment pairs is available \href{https://gitlab.com/lianeg/temporal-entailment-sports-dataset}{\color{blue} \myul[blue] {here}}.

\begin{table}[!htb]
    \begin{subtable}{0.5\linewidth}
      \centering
        \caption{1 = entailment, 0 = non-entailment. Blue = base (entailments, and non-entailments from temporally disjoint outcomes), orange = directional non-entailment, green = paraphrases}
        \begin{tabular}{|c|cccc|}
        \hline
        & win & lose & tie & play \\
        \hline
        win & \cellcolor{green!25}1 & \cellcolor{blue!25}0 & \cellcolor{blue!25}0 & \cellcolor{blue!25}1 \\
        lose & \cellcolor{blue!25}0 & \cellcolor{green!25}1 & \cellcolor{blue!25}0 & \cellcolor{blue!25}1 \\
        tie & \cellcolor{blue!25}0 & \cellcolor{blue!25}0 & \cellcolor{green!25}1 & \cellcolor{blue!25}1 \\
        play & \cellcolor{orange!25}0 & \cellcolor{orange!25}0 & \cellcolor{orange!25}0 & \cellcolor{green!25}1 \\
        \hline
        \end{tabular}
    \end{subtable}
    \begin{subtable}{0.5\linewidth}
      \centering
        \caption{Examples from the evaluation dataset}
        \begin{tabular}{|c|c|c|}
        \hline
        \textbf{Category} & \textbf{Examples} & \textbf{Size} \\
        \hline
        \cellcolor{blue!25} & defeat $\rightarrow$ vs & \\
        \multirow{-2}{*}{\cellcolor{blue!25}entailment 1} & crush $\rightarrow$ face & \multirow{-2}{*}{272}\\
        \hline
        \cellcolor{blue!25} & beat $\rightarrow$ fall to & \\
        \multirow{-2}{*}{\cellcolor{blue!25}outcome 0} & outscore $\rightarrow$ lose & \multirow{-2}{*}{446}\\
        \hline
        \cellcolor{orange!25} & play $\rightarrow$ win & \\
        \multirow{-2}{*}{\cellcolor{orange!25}directional 0} & go against $\rightarrow$ tie & \multirow{-2}{*}{272} \\
        \hline
        \cellcolor{green!25} & top $\rightarrow$ knock off & \\
        \multirow{-2}{*}{\cellcolor{green!25}paraphrase 1} & defeat $\rightarrow$ outplay & \multirow{-2}{*}{322} \\
        \hline
        \end{tabular}
    \end{subtable}
    \caption{Entailment pairs evaluation dataset}
    \label{tab:evaldataset}
\end{table}

\subsection{Similarity Measures}
\label{subsec:similarity_measures}
We compute both symmetric and directional similarity measures to learn entailments, making use of the temporally filtered counts and PMI scores described in Section~\ref{subsec:graph_construction}. Specifically, we adapted Lin's similarity measure \cite{lin1998}, Weeds' precision and recall measures \cite{weeds2003}, and BInc \cite{szpektor2008}. The adaptations of these measures are:

\textbf{Temporal count-based measures} using the temporally filtered counts: Weeds' precision, recall, and similarity (harmonic average of precision and recall); Lin's similarity; BInc using Weed's precision and count-based Lin's similarity.

\textbf{Temporal PMI-based measures:}
As a proxy to computing Conditional PMI between an entity pair, predicate \textit{p}, and predicate \textit{q}, which would be computationally expensive (if not infeasible) given the existing graph construction framework, we scale the original PMI scores. We apply two strategies: 1) \textit{Ratio:} scale the original PMI scores according to the ratio of filtered counts to regular counts, 2) \textit{Binary:} use the original PMI score if any of the events in predicate \textit{p} overlap with any event in predicate \textit{q}, otherwise set the score to zero. The following measures use the Ratio and Binary PMI strategies: Weeds' precision, recall, and similarity; Lin's similarity; BInc using Weed's PMI precision and Lin's similarity.

\textbf{Temporal hybrid BInc measures:} Computed using count-based Weeds' precision and PMI-based Lin's similarity, using the temporally filtered counts. We do this for both Ratio and Binary PMI.

We also ensure that for every temporal measure, its non-temporal counterpart is also included, and we include cosine similarity as a symmetric baseline. In total, we experiment with 29 similarity measures.

\section{Experiments, Results and Analysis}
\label{sec:results}

\subsection{Experimental Settings}
\label{sec:experiments}
As described in Section~\ref{subsec:relation_extraction} we extract all possible relations from the NewsSpike corpus and map their entities to types using the FIGER hierarchy. We construct a typed entailment graph for the organization/organization type pair using the subset of these relations where both entities are sports teams. We compute entailment scores using the set of 29 similarity measures described in Section~\ref{subsec:similarity_measures}. Due to space constraints we highlight results for eight of these measures: BInc and cosine similarity baselines, and the best performing temporal measures and their non-temporal counterparts. 

We experimented with different values for the event-specific time information source and temporal window described in Section~\ref{subsec:graph_construction}. We constructed typed entailment graphs using only time expressions (timexOnly), only the document creation date (docDateOnly), and using time expressions where available, otherwise backing off to the document creation date (timexAndDocDate). For each of the time interval sources, we also applied windows of 1, 2, 3, 4, 5, 6, 7, 30 and 3,650 days, as well as \textit{no window}.

We used the evaluation dataset described in Section~\ref{subsec:dataset} and Table~\ref{tab:evaldataset} to evaluate entailments captured under each of these experimental settings. We evaluate on three different configurations of the dataset: \textbf{Base} (entailment 1 and outcome 0), \textbf{Directional} (entailment 1 and directional 0), and \textbf{All} (entailment 1, outcome 0, directional 0, paraphrase 1). When we introduce parameter tuning in future work, we plan to assign a development/test split to the dataset.

\subsection{Temporal Information Source}
\label{sec:results_timesource}
See Table~\ref{tab:time_source} for area under the curve (AUC) scores for each of the three temporal information sources, for a range of temporal (T) and non-temporal similarity measures. To evaluate the similarity measures fairly we calculate AUC under a recall threshold (a recall of 0.75 is reached by all non-timexOnly measures). The \textit{timexOnly} source has the weakest performance, since it has access to a sparser set of time intervals (with only 19\% of relations linked to a time expression in the text). When we focus on the low recall range ($<$ 0.15), however, we find that the temporal measures outperform the non-temporal ones. This is especially promising as it highlights the benefit of using the more accurate time intervals resolved from time expressions in the text; these give temporal measures a larger increase than the document date. When using only the document creation date (\textit{docDateOnly}), we find that temporal and non temporal measures perform similarly. Results for the \textit{timexAndDocDate} source show that leveraging both time expression and document time information together leads to the most effective model.

\begin{table}[]
\centering
\small
\begin{tabular}{lcccc}
\toprule
\textbf{Similarity measure}                            & \multicolumn{2}{c}{\textbf{timexOnly}} & \textbf{docDateOnly} & \textbf{timexAndDocDate} \\ \midrule
                   & rec $<$ 0.15               & rec $<$ 0.75               & rec $<$ 0.75                 & rec $<$ 0.75                     \\ 
\midrule
T. BInc (Count)             & 0.111              & 0.127              & \textbf{0.467}       & \textbf{0.469}           \\ 
BInc (Count)                & 0.106              & \textbf{0.467}     & \textbf{0.467}       & 0.467                    \\ 
T. BInc (Ratio PMI)         & 0.108              & 0.108              & 0.446                & 0.449                    \\ 
T. BInc (Binary PMI)        & \textbf{0.114}     & 0.114              & 0.446                & 0.448                    \\
BInc                        & 0.093              & 0.444              & 0.444                & 0.444                    \\
T. Weed's Pr (Count)         & 0.113              & 0.129              & 0.436                & 0.436                    \\
Weed's Pr (Count)            & 0.087              & 0.436              & 0.436                & 0.436                    \\
Cosine Sim                  & 0.100              & 0.420              & 0.420                & 0.420  \\   \bottomrule
\end{tabular}
\caption{Temporal information source: AUC scores for the base evaluation dataset, and a temporal window size of 4 days}
\label{tab:time_source}
\end{table}

\subsection{Temporal Window Size}
\label{sec:results_timewindow}
\begin{figure}[h]
\begin{center}
\includegraphics[width=12.5cm]{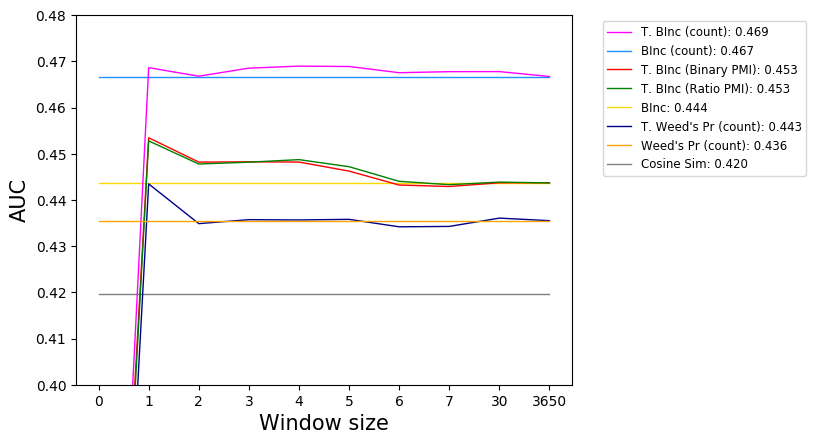}
\end{center}
\caption{Effects of window size for the \textit{timexAndDocDate} temporal information source}
\label{fig:window_size}
\end{figure}

In Figure~\ref{fig:window_size} we can see that there is a sharp improvement in AUC score for all of the temporally-informed similarity measures when a window of one day is applied\footnote{With no window, the temporally informed similarity measures perform poorly (between 0.24 and 0.26)}. This is likely due to data sparsity and because sports articles report on the same event on different days. The horizontal lines represent the non-temporally-informed similarity measures. We can also see that the choice of window size depends on the similarity measure. For the majority of the temporally-informed similarity measures, a window size between one and four days works well. For this class of predicates a window size of 4 seems suitable, as it avoids conflating games that happen on consecutive weekends, while giving some leeway. We discuss the possibility of using a dynamic window in Section \ref{sec:discussion}.

\subsection{Comparing Similarity Measures}
\label{sec:results_simmeasures}

\begin{figure}[h]
\begin{center}
\includegraphics[width=10cm]{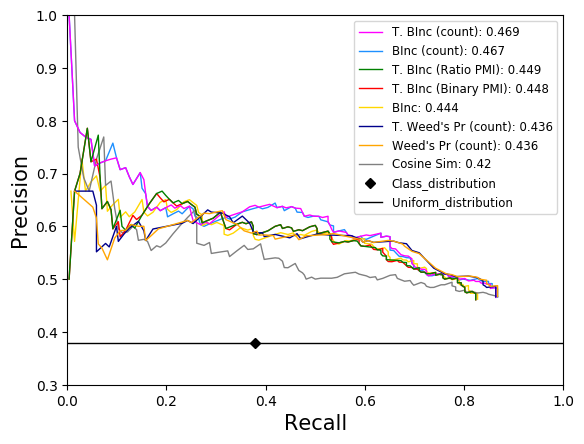}
\end{center}
\caption{Best results on the base evaluation dataset: timexAndDocDate with a temporal window size of 4 days. For calculating AUC the recall threshold is set to $<$ 0.75 for all similarity measures}
\label{fig:best_graph}
\end{figure}

We can see from Figure~\ref{fig:best_graph} that \textit{BInc}, a state-of-the-art measure for relation entailment, does not perform well within the temporal setting and is outperformed by a number of the temporal measures. Of those, \textit{T. BInc count sims}, the version of BInc that uses the temporally filtered counts, produces the best results on the \textit{base} subset of the evaluation dataset. The two temporally informed measures that use scaled PMI scores (\textit{T. Ratio BInc sims} and \textit{T. Binary BInc sims}) also outperform \textit{BInc} but to a lesser degree. This may be due to sparsity in the set of binary relations making it difficult to estimate accurate scaled PMI scores. We hope to alleviate this problem by moving to a larger corpus in future work\footnote{We have already compiled a dataset of 10 years worth of news data from multiple sources}. In more general terms, for each similarity measure, the temporal version performs better (with the exception of Weed's probabilistic precision, for which there is no change).

\subsection{Performance on Data Subsets}
\label{sec:results_data_subsets}
Our main interest is in performance on the \textit{base} dataset, but for completeness and comparison to previous work, which tests directionality more, we also evaluated on the \textit{all} and \textit{directional} subsets (see Table~\ref{tab:subsets}).

To investigate the challenge of directionality in entailment we consider the set of entailments and their reverse, e.g. play $\rightarrow$ win (``entailment 1" and ``directional 0" in Table~\ref{tab:evaldataset}(b)). We find that in general the temporal similarity measures still perform strongly. \textit{T. Weeds' precision}, the only purely directional measure in the set, performs comparably to its non-temporal counterpart on the \textit{directional} subset. The second best measure is \textit{BInc} which is unsurprising given that it also captures directionality, and that the dataset no longer tests temporality.

We also evaluate on the complete dataset (\textit{all}) which includes paraphrases (``paraphrase 1" in Table~\ref{tab:evaldataset}(b)). Here we find that \textit{T. Weeds' precision} is again the best measure, closely followed by the PMI-based \textit{T. BInc} scores (Binary and Ratio). We expect that the strong performance of \textit{T. Weeds' precision} is due to correctly identifying the directional entailments in the dataset, and that the drop in performance between the \textit{directional} and \textit{all} subsets is due to the inclusion of paraphrases, which are challenging for all measures, but particularly for the Weed's precision measures which have no symmetric component. \textit{BInc} also performs reasonably well on this subset, showing that temporally-uninformed measures remain competitive when multiple phenomena are tested. In general we find that at least one temporally-informed similarity measure still performs strongly for each of the subsets in the dataset.

\begin{table}
\centering
\begin{tabular}{lccc}
    \toprule
    \textbf{Similarity measure}          & \textbf{Base}    & \textbf{Dir} & \textbf{All}     \\
    \midrule
    T. BInc (Count)             & \textbf{0.469} & 0.433                & 0.433          \\ 
    BInc (Count)                & 0.467          & 0.433                & 0.434          \\ 
    T. BInc (Ratio PMI)         & 0.449          & 0.449                & 0.442          \\ 
    T. BInc (Binary PMI)        & 0.448          & 0.449                & 0.442          \\ 
    BInc                        & 0.444          & 0.447                & 0.442          \\ 
    T. Weed's Pr (Count)         & 0.436          & 0.491                & \textbf{0.448} \\ 
    Weed's Pr (Count)            & 0.436          & \textbf{0.492}       & 0.447          \\ 
    Cosine Sim                  & 0.420          & 0.372                & 0.397          \\
    \bottomrule
    \end{tabular}
    \captionof{table}{AUC scores for different subsets of the evaluation dataset. Temporal information source is timexAndDocDate, temporal window size is 4 days}
    \label{tab:subsets}
\end{table}

\subsection{Effects of Using Temporal Information}
\label{sec:results_best}
To see the benefits of including temporal information, we compare the AUC scores for the temporal and non-temporal similarity measures. We use the \textit{timexAndDocDate} temporal information source, and a window size of four days (see Figure~\ref{fig:best_graph}). The organization-organization graph has 46,899 nodes and between 7,888,372 and 8,256,182 edges depending on the similarity measure. We can see that the temporal similarity measure \textit{T. BInc count} outperforms \textit{BInc}, the state-of-the-art non-temporal similarity measure for relation entailment employed by \newcite{berant2011} and \newcite{hosseini2018}. We conclude that incorporating temporal information is beneficial for accurately predicting entailments and non-entailments for highly correlated predicates that are part of distinct events.

\section{Discussion and Future Work}
\label{sec:discussion}

The NewsSpike corpus covers a period of approximately six weeks so the outcomes of two matches between two teams within this period may be fairly similar (as there will have been few changes to the teams management, players, etc.). In future work we plan to move to a corpus covering a larger time period, for which we would expect to observe a greater effect. Expanding beyond the sports team domain would also allow us to study events with longer duration, such as a president holding office, preceded by their campaign, and election. 

Expanding the domain will also give us a collection of local graphs of types beyond just organization-organization, across which we can then learn globally consistent similarity scores (as in \newcite{hosseini2018}). We can then also collapse cliques in the graph into paraphrase clusters with a single relation identifier, which we hope will improve performance, especially on sparser predicates. Our dataset is perfectly suited to evaluating benefits from this addition due to its origin in paraphrase clusters.

Whilst we can observe a positive effect when using temporal information, the effect is modest. Upon closer inspection we found that this was due to relatively few events being filtered. An analysis of a subset of sentences revealed that relations were being extracted spuriously due to various linguistic phenomena. Issues are caused by conditionals (e.g. ``if Arsenal win”), modals (``I still expect Arsenal…”), incorrect future predictions (``Arsenal will win”) and counterfactuals (``had Arsenal won,...”). These types of predictions appear to be especially common in the sports domain. Another issue arose due to an incorrect application of passive to active conversion (\textbf{Arsenal}-\textit{lost to}-\textbf{Man United} from ``Man United lost to Arsenal") resulting from incorrect verb feature labels in the CCG parses. Finally, SUTime sometimes provides partial time information which can result in a whole year being used as an interval, creating spurious overlaps\footnote{Focusing only on short time intervals failed to offset the other sources of spurious overlaps}. Addressing these issues should lead to a larger effect from using temporal information, because it would reduce overlaps and allow more filtering.

Our algorithm lends itself naturally to mining entailments with a temporal relation such as \textit{visit} $\rightarrow$ \textit{arrived}. We plan to achieve this by splitting the window into a before and after frame, producing separate entailment scores for different orderings. We also plan to investigate setting the window dynamically. In the current setup, events stay relevant for a similar amount of time, but different predicates should allow comparison for different granularities of time. For example, the window around a person \textit{being president} should be larger than a person \textit{visiting} a location. Initially we might aim to learn a different window size per predicate (for example by taking into account average predicate duration and granularity). Dynamic windowing could become particularly valuable with a broader domain.

More generally, our method could incorporate in its filtering any function of the contextualised events to determine whether their co-occurrence should contribute to an entailment score. Currently a binary decision is made based on time interval overlap, but one might use features such as (lexical) aspect, tense, the presence of other entities, etc. Previous work was limited to using the presence of two entities as a proxy for entailment relevance; with our refinements we could expand to involving not only time but also other features of the contextualised events.

We will also explore the use of other temporal resolution systems and aim to develop more sophisticated ways of linking times to events, which currently only occurs through CCG dependencies. More time intervals might be propagated using the TempEval \cite{uzzaman2013semeval} ordering approach or through other means, for instance by reasoning about tense, Reichenbachian reference time \cite{reichenbach1947tenses}, or event coreference (within, or across documents).

\section{Conclusions}
\label{sec:conclusions}
We injected temporal information into the local entailment graph construction method of \newcite{hosseini2018}, with the goal of comparing only those events that are temporally near each other. This is achieved by filtering the counts of predicate \textit{p} according to whether its events' time intervals overlap with the those of predicate \textit{q}. We considered a range of new local similarity scores based on both temporally filtered counts and scaled PMI scores, which we evaluate on a semi-automatically constructed dataset, based on manually constructed paraphrase clusters. 

Our temporal similarity measures outperform their non-temporal counterparts, including BInc, the state-of-the-art measure for relation entailment. We also show that using a combination of time expressions recovered from the text and the document creation date performed better than using only one of these sources, and that adding a temporal window around the time intervals of the events is essential. The performance of the temporal similarity measures over the non-temporal measures is particularly strong at the low recall range when only time expressions from the text are used. This is especially promising as it suggests that there is much room for improvement in using more sophisticated temporal resolution systems and methods for linking times to events.

\section*{Acknowledgements}
\label{sec:acknowledgements}

We thank Miloš Stanojević for assistance with the Rotating CCG parser, Nick McKenna and Ian Wood for helpful discussions, and the reviewers for their valuable feedback. This work was funded by the ERC H2020 Advanced Fellowship GA 742137 SEMANTAX, a grant from The University of Edinburgh and Huawei Technologies, and The Alan Turing Institute under EPSRC grant EP/N510129/1.

\bibliographystyle{coling}
\bibliography{coling2020}

\end{document}